\documentclass[11pt,a4paper]{article}

\usepackage{amsmath,amssymb,amsfonts} 
\usepackage{algorithmic} 
\usepackage{graphicx} 
\usepackage{booktabs} 
\usepackage{hyperref} 
\usepackage{listings} 
\usepackage{float} 
\usepackage{geometry} 
\usepackage{tikz}
\usepackage{authblk} 

\usetikzlibrary{positioning, shapes.geometric, arrows.meta}
\title{From Spectra to Geography: Intelligent Mapping of RRUFF Mineral Data}
\author[1,2]{Francesco Pappone}
\author[3]{Federico Califano}
\author[2]{Marco Tafani}

\affil[1]{PSTP Technoscience, Centro di Competenza per L'Intelligenza Artificiale Applicata}
\affil[2]{Dipartimento di Medicina Sperimentale, Sapienza Università di Roma}
\affil[3]{Dipartimento di Ingegneria Meccanica e Aerospaziale, Sapienza Università di Roma}

\date{\today}

\begin{document}

\maketitle

\begin{abstract}
Accurately determining the geographic origin of mineral samples is pivotal for applications in geology, mineralogy, and material science. Leveraging the comprehensive Raman spectral data from the RRUFF database, this study introduces a novel machine learning framework aimed at geolocating mineral specimens at the country level. We employ a one-dimensional ConvNeXt1D neural network architecture to classify mineral spectra based solely on their spectral signatures. The processed dataset comprises over 32,900 mineral samples, predominantly natural, spanning 101 countries. Through five-fold cross-validation, the ConvNeXt1D model achieved an impressive average classification accuracy of 93\%, demonstrating its efficacy in capturing geospatial patterns inherent in Raman spectra. 

\end{abstract}

\tableofcontents

\section{Introduction}

The geolocation of mineral samples is essential for various applications in geology, mineralogy, and material science. Traditional methods rely on manual and qualitative identification, as well as on contextual information. Such approaches can be time-consuming and subjective, creating a bottleneck effect in reliable characterization of minerals. With the increasing availability of spectral databases \cite{el2019raman,MATTIODA2024115769, berlanga2024mlrod}, there is an opportunity to leverage machine learning techniques to automate the geolocation process. Previous studies have applied machine learning to spectral data for mineral classification \cite{berlanga2022convolutional, bendinelli2024gemtelligence}, yet there is limited research on utilizing spectral data specifically for geolocation. A notable approach, limited to gemstones, has been carried out by Gemintelligence \cite{bendinelli2023gemtelligenceacceleratinggemstoneclassification}. Previous work has specifically focused on applying convolutional neural networks and architectural variants of ResNets \cite{he2015deepresiduallearningimage,LafuenteDownsYangStone+2016+1+30} to Raman Spectroscopy,  To bridge this gap, this paper explores the use of a one-dimensional ConvNeXt neural network \cite{liu2022convnet2020s}, referred to as ConvNeXt1D, which to our knowledge has not been previously applied in Raman spectral deep learning. ConvNeXt1D is employed here to classify Raman spectra of minerals for geolocation purposes, focusing on predicting the country of origin based on spectral characteristics. The RRUFF Project \cite{LafuenteDownsYangStone+2016+1+30} provides an integrated database of Raman spectra, X-ray diffraction, and chemistry data for minerals, serving as a valuable resource for the identification and characterization of mineral samples. In this work, we extended the RRUFF database by geocoding the locality information associated with each mineral sample, converting textual locality descriptions into geographic coordinates (latitude and longitude). This geographic information is essential for our machine learning objectives, allowing the analysis of spatial patterns and the development of location-based predictive models.

\section{Data Processing and Cleaning}

To prepare the dataset for analysis, several data processing and cleaning steps were undertaken to ensure its reliability for subsequent examination. Each mineral sample in the RRUFF database is associated with a locality description containing geographic information. To extract precise geographic coordinates from these descriptions, geocoding services such as Nominatim \cite{Nominatim}, Photon \cite{Photon}, and the ArcGIS Geocoding Service \cite{Esri} were employed. The geocoding process began with cleaning the locality strings by removing any text within parentheses, eliminating extra spaces, and standardizing delimiters. These cleaned strings were then sequentially submitted to each geocoder until valid latitude and longitude coordinates were obtained. In instances where the full locality string could not be geocoded, partial matches were attempted by progressively removing segments of the string from the beginning. Despite these efforts, some localities remained un-geocoded due to missing or ambiguous information.




The dataset comprises both natural and synthetic mineral samples. To differentiate between them, specific keywords indicative of synthetic origins were searched within the \texttt{locality} and \texttt{names} fields. Samples containing any of these keywords were classified as synthetic and were treated accordingly during the analysis process.
Addressing missing data was critical to maintain the dataset's integrity. Samples lacking latitude or longitude values after the geocoding process were retained but flagged, and subsequently excluded from analyses that required geographic information. Additionally, samples with incomplete metadata were carefully reviewed. If essential information, such as the mineral name, was missing, the sample was excluded from the dataset. 



To associate each geocoded sample with a country, a spatial join was performed using a country shapefile from Natural Earth \cite{NaturalEarth}. The geocoded points were spatially joined with the country polygons to assign a country name to each sample. This process enabled the analysis of the distribution of mineral samples by country and facilitated the production of visualizations such as choropleth maps.

\section{Dataset Overview and Statistics}
The processed RRUFF dataset comprises a comprehensive collection of mineral samples, providing a robust foundation for our analysis. In total, the dataset includes \textbf{32,940} samples, with the vast majority being natural samples (32,214 samples, or \textbf{97.80\%}) and a smaller proportion of synthetic samples (726 samples, or \textbf{2.20\%}). This is advantageous for analyses focusing on naturally occurring minerals. Geographically, the dataset is highly complete, with 32,892 samples (99.85\%) containing both latitude and longitude coordinates, while only 48 samples (0.15\%) lack geographic information. The dataset showcases an impressive diversity of minerals, encompassing a total of 2,027 unique mineral species. This extensive variety is particularly beneficial for the development of machine learning models, as it provides a wide range of mineral characteristics for training and evaluation.

\begin{figure}[H]
\centering
\includegraphics[width=\textwidth]{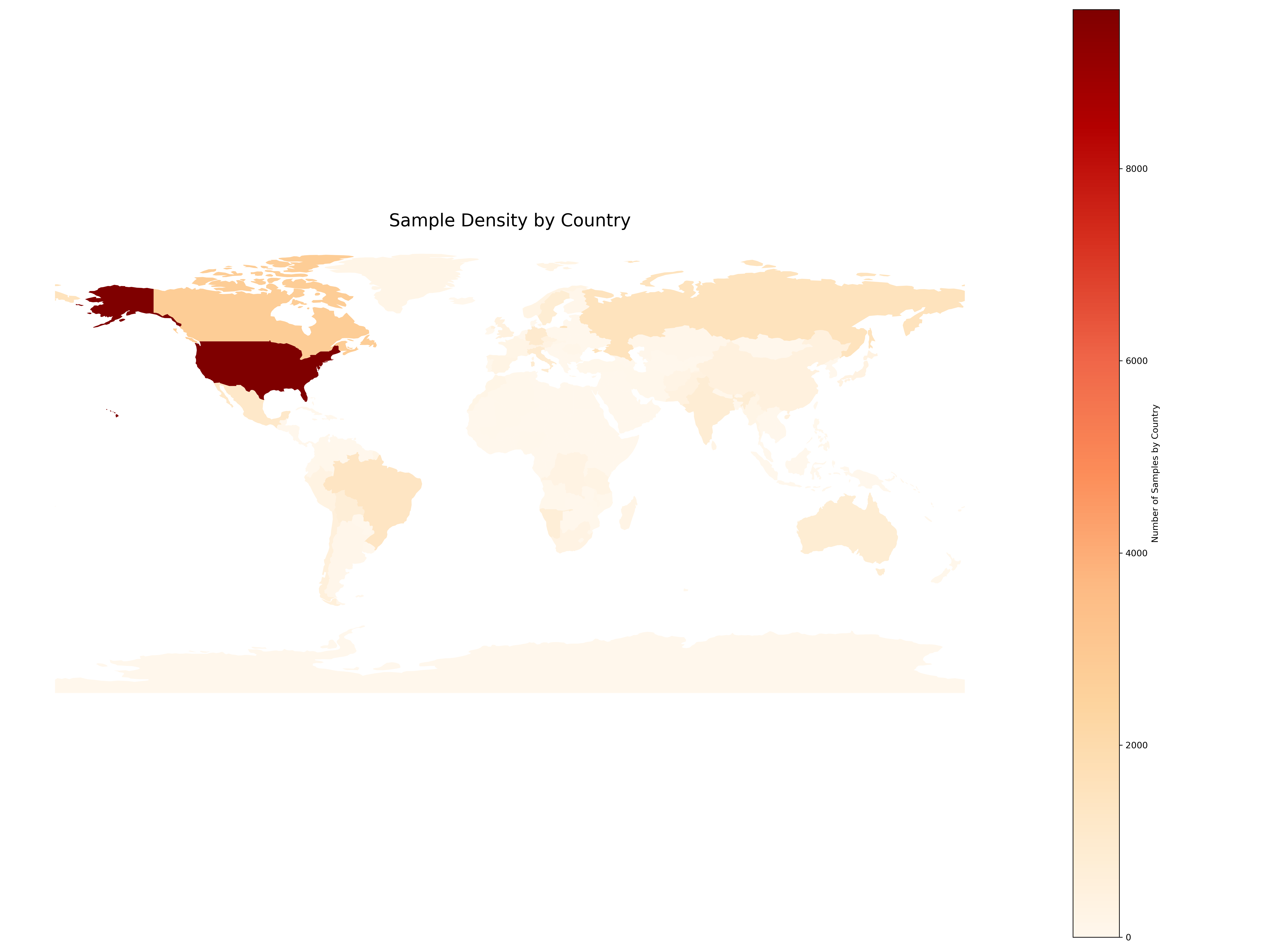}
\caption{Sample Density by Country}
\label{fig:sample_density_map}
\end{figure}

Understanding the spatial distribution of the samples is crucial for identifying potential regional biases and assessing the representativeness of the data. The geographic distribution analysis reveals the number of natural samples per country, as illustrated in Figure~\ref{fig:sample_density_map} through a choropleth map. This visualization highlights the concentration of mineral samples across different countries, enabling insights into regional data density and facilitating targeted analyses based on geographic trends.





\begin{figure}[H]
\centering
\includegraphics[width=\textwidth]{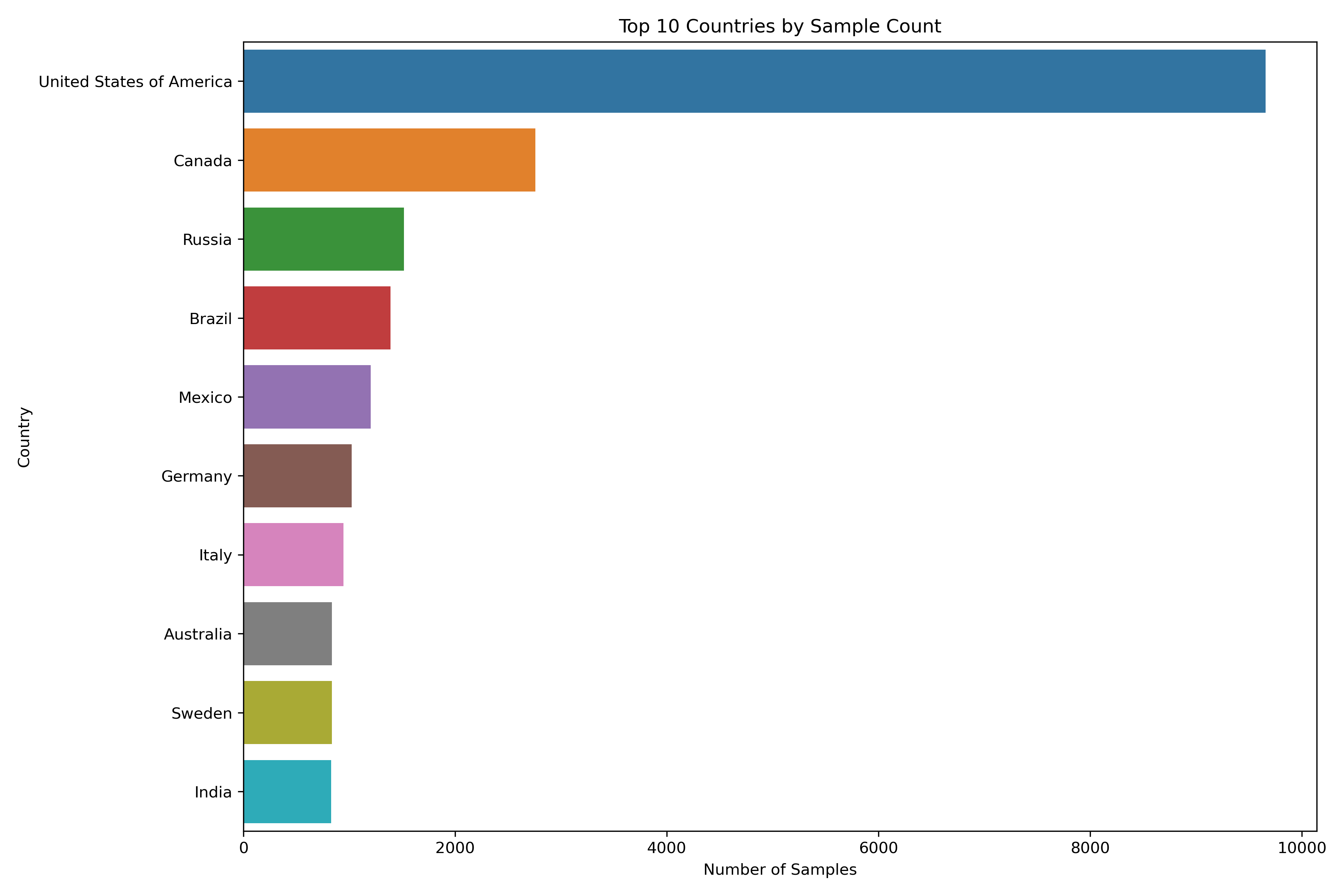}
\caption{Top 20 Countries by Sample Count}
\label{fig:top_countries}
\end{figure}

\begin{figure}[H]
\centering
\includegraphics[width=\textwidth]{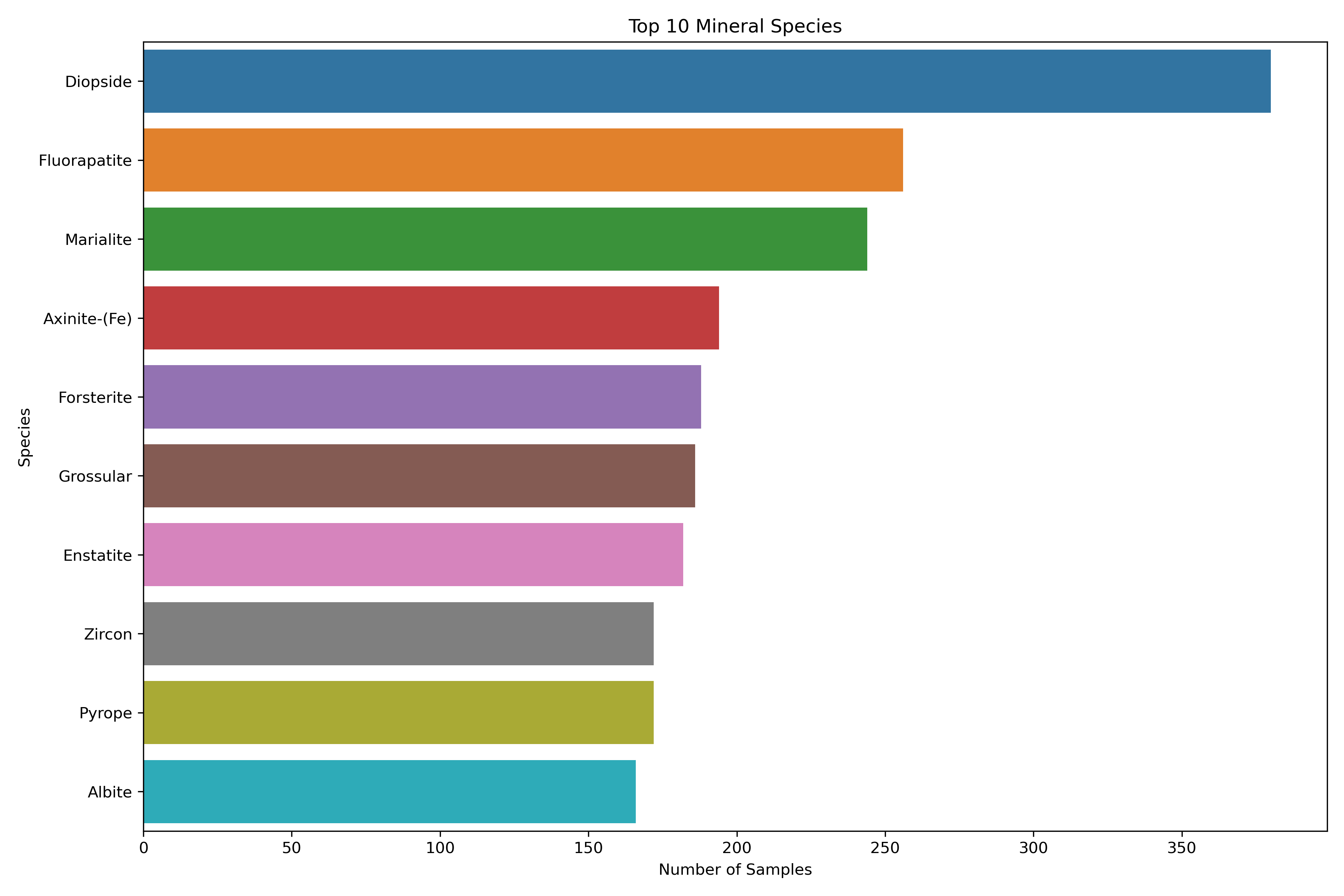}
\caption{Top 20 Mineral Species in the Dataset}
\label{fig:top_species}
\end{figure}

The map underscores regions with higher concentrations of samples, with the United States of America having the highest number of samples (9,656), accounting for approximately 30.96\% of the dataset. Other countries with significant sample counts include Canada, Russia, Brazil, and Mexico. These concentrations indicate areas of extensive mineral exploration and collection, which may influence the dataset's overall representativeness.
This distribution suggests a concentration of data in certain regions, which may influence the generalizability of models trained on this dataset.
The top 10 countries by sample count are presented in Figure~\ref{fig:top_countries} and the top 20 detailed in Table~\ref{tab:top_countries}. The top 4 countries contribute to over 50\% of the total samples, indicating a potential geographic bias toward these regions. Figure~\ref{fig:top_species} displays the top 10 mineral species in the dataset, and Table~\ref{tab:top_species} provides sample counts for the top 20. The presence of a wide variety of mineral species enhances the dataset's utility for training machine learning models with broad applicability.

\begin{table}[H]
\centering
\caption{Counts and Cumulative Percentages of Top Countries}
\label{tab:top_countries}
\begin{tabular}{clcc}
\toprule
\textbf{Rank} & \textbf{Country}              & \textbf{Sample Count} & \textbf{Cumulative Percentage (\%)} \\
\midrule
1  & United States of America & 9,656  & 30.96 \\
2  & Canada                   & 2,758  & 39.80 \\
3  & Russia                   & 1,514  & 44.66 \\
4  & Brazil                   & 1,390  & 49.12 \\
5  & Mexico                   & 1,202  & 52.97 \\
6  & Germany                  & 1,022  & 56.25 \\
7  & Italy                    & 946    & 59.28 \\
8  & Australia                & 836    & 61.96 \\
9  & Sweden                   & 834    & 64.63 \\
10 & India                    & 828    & 67.29 \\
11 & Namibia                  & 754    & 69.71 \\
12 & Bolivia                  & 730    & 72.05 \\
13 & Chile                    & 582    & 73.91 \\
14 & United Kingdom           & 466    & 75.41 \\
15 & China                    & 462    & 76.89 \\
16 & Czechia                  & 446    & 78.32 \\
17 & Pakistan                 & 432    & 79.70 \\
18 & Japan                    & 390    & 80.95 \\
19 & Peru                     & 348    & 82.07 \\
20 & Dem. Rep. Congo          & 332    &83.13\\
\bottomrule
\end{tabular}
\end{table}

\begin{figure}[H]
    \centering
    \includegraphics[width=0.8\textwidth]{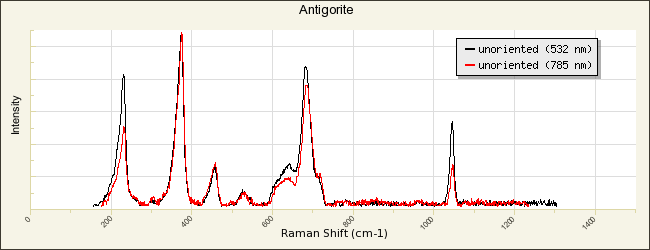}
    \caption{A pair of example spectra from the RRUFF dataset}
    \label{fig:image_from_RRUFF}
\end{figure}

\begin{table}[H]
\centering
\caption{Counts of Top Mineral Species}
\label{tab:top_species}
\begin{tabular}{clc}
\toprule
\textbf{Rank} & \textbf{Mineral Species} & \textbf{Sample Count} \\
\midrule
1  & Diopside      & 380 \\
2  & Fluorapatite  & 256 \\
3  & Marialite     & 244 \\
4  & Axinite-(Fe)  & 194 \\
5  & Forsterite    & 188 \\
6  & Grossular     & 186 \\
7  & Enstatite     & 182 \\
8  & Zircon        & 172 \\
9  & Pyrope        & 172 \\
10 & Albite        & 166 \\
11 & Actinolite    & 162 \\
12 & Beryl         & 162 \\
13 & Kyanite       & 156 \\
14 & Epidote       & 150 \\
15 & Microcline    & 146 \\
16 & Titanite      & 140 \\
17 & Tremolite     & 140 \\
18 & Elbaite       & 134 \\
19 & Andalusite    & 132 \\
20 & Rutile       & 132 \\
\bottomrule
\end{tabular}
\end{table}
The dataset has been extensively utilized by the community as a benchmarking standard, especially for machine learning and deep learning techniques \cite{spectral_matching,unified_raman}, due to its high volume of diverse spectra, acquired via a variety of different instruments and under diverse conditions. The high percentage of natural samples and extensive species diversity are particularly beneficial, providing a rich and varied dataset that supports robust model training and evaluation. An example of spectra can be seen in \ref{fig:image_from_RRUFF}.
However, certain limitations must be acknowledged to contextualize the dataset's applicability. Firstly, there is a \textbf{geographic bias} due to the concentration of samples in a few countries. This uneven distribution may introduce bias, potentially skewing analyses toward the geological characteristics prevalent in these regions. Secondly, some regions are \textbf{underrepresented}, with sparse data that could affect model performance and generalizability in those areas. These limitations highlight the need for cautious interpretation of the results and suggest avenues for future data collection efforts to achieve a more balanced and comprehensive dataset.





\section{Methodology}

\subsection{Spectral Data Processing}
We processed the spectral data to find a wavenumber window covering the entirety of the dataset. We then padded spectra in regions that were not represented in the original spectrum with zeros. The intensities were normalized via a MinMax approach, and the spectra were interpolated utilizing cubic splines, and then resampled on homogenous grids to ensure uniform input dimensions for the neural network. Further details regarding the preprocessing pipeline can be found in appendix A. 

\begin{figure}[H]
    \centering
    \includegraphics[width=0.8\textwidth]{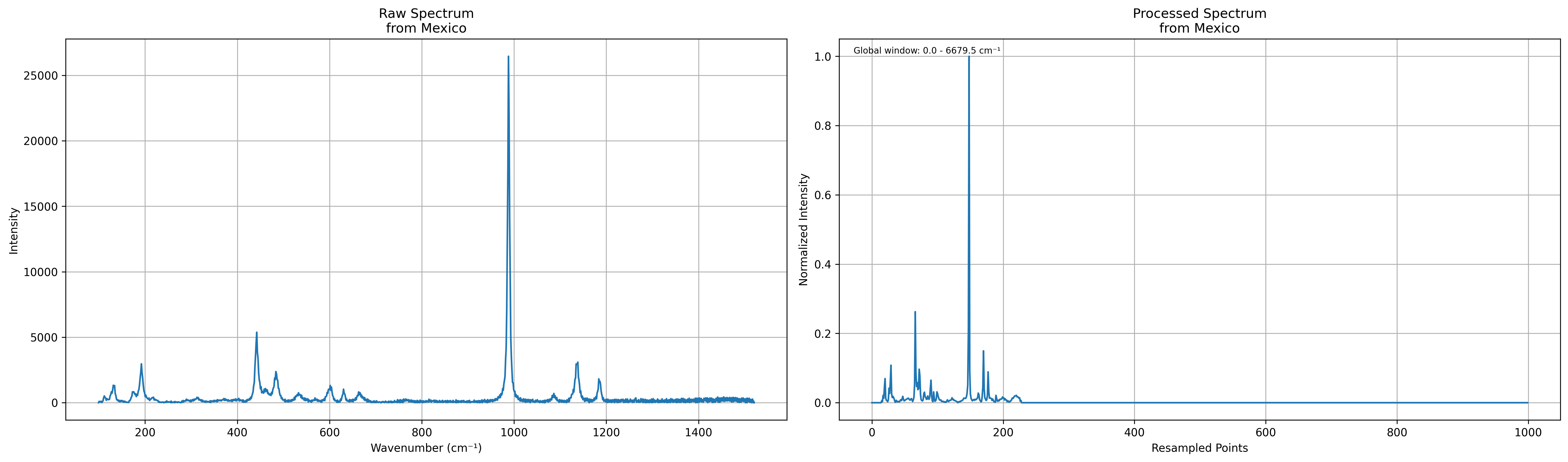}
    \caption{An example of a spectrum from Mexico being processed. On the left, the spectrum is raw, while on the right we can see the reinterpolated version over a much larger window, which is appropriately selected to cover the entirety of the dataset and padded to represent the spectrum in the correct window.}
    \label{fig:processingofspectra}
\end{figure}

\subsection{Model Architecture}
The ConvNeXt architecture is a pure convolutional neural network that aims to bridge the gap between traditional ConvNets and vision Transformers \cite{liu2022convnet2020s}. It is the result of a systematic modernization of the ResNet architecture, incorporating design choices inspired by vision Transformers while maintaining the simplicity and efficiency of standard ConvNets.
Liu et al. started with a standard ResNet and gradually transformed it, adopting various architectural innovations from vision Transformers. This process led to a family of models (ConvNeXt-T/S/B/L/XL) that compete favorably with state-of-the-art hierarchical vision Transformers like Swin Transformer across various computer vision benchmarks.  The core of our classification system is a modified ConvNeXt architecture adapted for one-dimensional spectral data, which we call \textbf{ConvNeXt1D}. This architecture is designed to effectively process and classify Raman spectra, inheriting the strong shift-equivariant inductive bias of convolutional neural networks \cite{bronstein2021geometricdeeplearninggrids} while increasing in expressivity thanks to modern architectural tweaks.

\begin{figure}[H]
    \centering
    \includegraphics[width=0.8\textwidth]{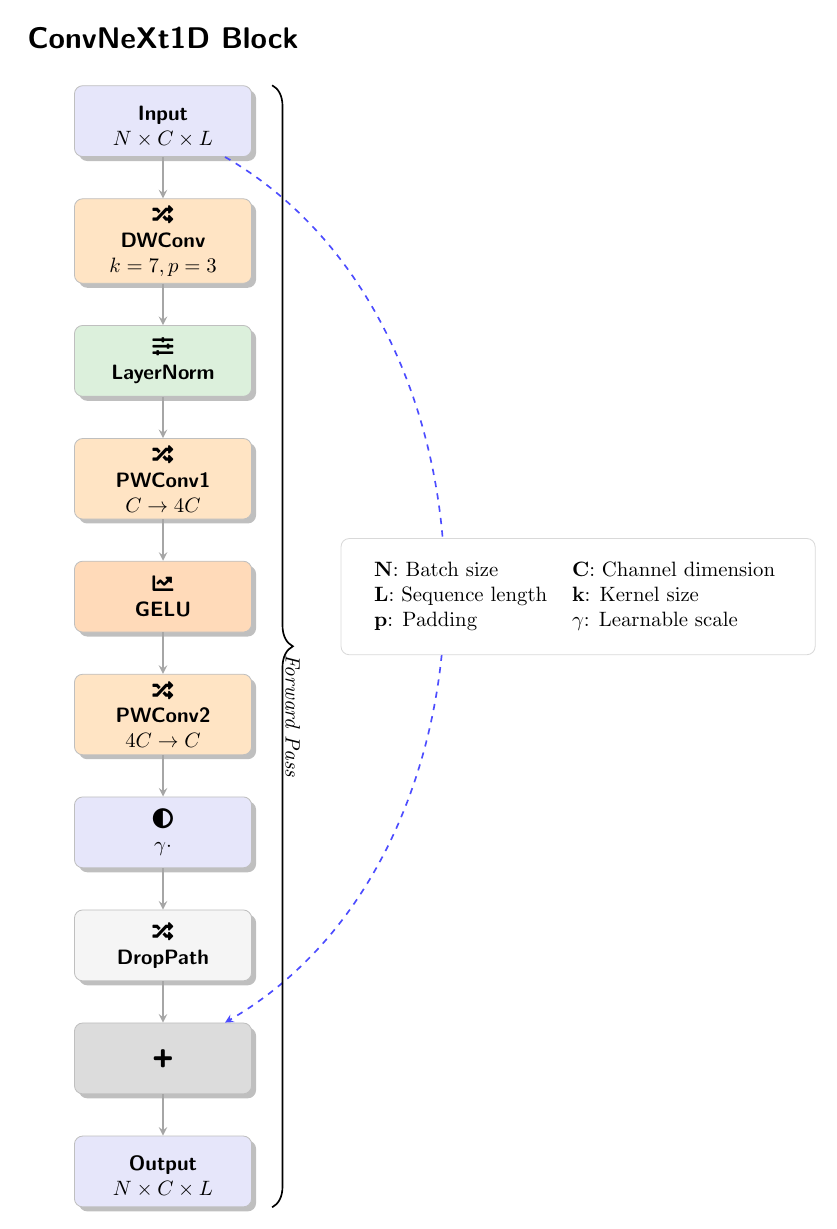}
    \caption{ConvNeXt1D Block architecture}
    \label{fig:convnext1d}
\end{figure}

\paragraph{Overall Structure}
The ConvNeXt1D model can be described as:
\begin{equation}
\text{ConvNeXt1D} = f_{\text{classifier}} \circ f_{\text{stages}} \circ f_{\text{stem}}
\end{equation}
where $f_{\text{stem}}$ is the input processing layer, $f_{\text{stages}}$ represents the main convolutional stages, and $f_{\text{classifier}}$ is the final classification layer.

\paragraph{Stem Layer}
The stem layer processes the input spectrum:
\begin{equation}
f_{\text{stem}}(x) = \text{LayerNorm}(\text{Conv1D}(x, \text{kernel\_size}=4, \text{stride}=4))
\end{equation}
This layer aggressively downsamples the input by a factor of 4, similar to the ``patchify'' stem in vision Transformers.

\paragraph{Main Stages}
The main part of the network consists of four stages, each comprising multiple ConvNeXt1D blocks:
\begin{equation}
f_{\text{stages}}(x) = f_{\text{stage}_4} \circ f_{\text{stage}_3} \circ f_{\text{stage}_2} \circ f_{\text{stage}_1}(x)
\end{equation}
Each stage $i$ is defined as:
\begin{equation}
f_{\text{stage}_i}(x) = f_{\text{block}_{N_i}} \circ \dots \circ f_{\text{block}_2} \circ f_{\text{block}_1}(x)
\end{equation}
where $N_i$ is the number of blocks in stage $i$, determined by the \texttt{depths} parameter.

\paragraph{ConvNeXt1D Block}
The ConvNeXt1D block serves as the foundational component of this architecture, drawing inspiration from the Transformer block while using solely convolutional operations. The block begins with a depthwise convolution, denoted as \(\text{DWConv}(x)\), applied to the input \(x\) with a kernel size of 7 and padding of 3. This depthwise convolution provides spatial filtering tailored for 1-dimensional data.
Next, layer normalization, written as \(\text{LayerNorm}(x)\) \cite{ba2016layernormalization}, is applied. This normalization technique is chosen over the traditional batch normalization, stabilizing activations independently across channels.
The normalized output then passes through a pointwise convolution, \(\text{PWConv1}(x)\), which acts as a linear layer expanding the feature dimension. This expansion phase aligns with the inverted bottleneck architecture commonly seen in Transformer models, where the dimension is briefly increased before subsequent processing.
Following this dimensional expansion, the \(\text{GELU}\) activation function, \(\text{GELU}(x)\), is applied \cite{hendrycks2023gaussianerrorlinearunits}. Finally, another pointwise convolution, denoted \(\text{PWConv2}(x)\), projects the expanded representation back to its original dimensionality.
At this point, a learnable parameter, \(\gamma\), scales the output, allowing for fine-tuning within each ConvNeXt1D block. To encourage model robustness, stochastic depth regularization, \(\text{DropPath}\), is then applied to the scaled output, which is subsequently added to the initial input \(x\) via a residual connection.

Thus, the complete block architecture can be summarized as follows:
\begin{equation}
y = x + \text{DropPath} \left( \gamma \cdot \text{PWConv2}\left(\text{GELU}\left(\text{PWConv1}\left(\text{LayerNorm}\left(\text{DWConv}(x)\right)\right)\right)\right) \right).
\end{equation}
\paragraph{Key Architectural Features}
In ConvNeXt1D, three architectural choices distinguish it from traditional convolutional networks. First, the use of grouped convolutions enables depthwise separable convolutions, reducing computational costs while maintaining performance. Second, the block’s inverted bottleneck structure, where the dimension expands and then projects back, mirrors the MLP block design of Transformers. Finally, the adoption of layer normalization in place of batch normalization allows ConvNeXt1D to stabilize activation distributions independently of batch size.

\paragraph{Hyperparameters}
The model's architecture is defined by two key hyperparameters:
\begin{itemize}
    \item \textbf{Depths}: $[2, 2, 3, 2]$, representing the number of ConvNeXt1D blocks in each stage.
    \item \textbf{Dimensions}: $[32, 64, 128, 256]$, representing the number of channels in each stage.
\end{itemize}

\paragraph{Classification Head}
After the final convolutional stage, we apply a global average pooling operation to aggregate the features across the temporal dimension. This operation reduces the dimensionality by computing the mean of the feature vectors along the sequence length $L$:

\begin{equation} x_{\text{pooled}} = \text{LayerNorm}\left( \frac{1}{L} \sum_{i=1}^{L} x_i \right) \end{equation}

where $x_i \in \mathbb{R}^C$ represents the feature vector at position $i$, $C$ is the number of channels, and $L$ is the sequence length. The Layer Normalization ensures that the pooled features are properly scaled before classification.
The final classification layer is then a simple linear projection applied to the pooled and normalized features.
gin of the input Raman spectrum.

\subsection{Training Procedure}
We split the dataset into training and testing sets with an 80-20 ratio. The model was trained using the ScheduleFree AdamW optimizer \cite{defazio2024roadscheduled} and a learning rate of 1e-3.

\paragraph{Model Initialization}
The model parameters are initialized using a truncated normal distribution:
\begin{equation}
w \sim \mathcal{TN}(0, 0.02^2)
\end{equation}
Biases are initialized to zero, and the layer scale parameter $\gamma$ is initialized to $10^{-6}$.

\paragraph{Training Objective}
We use the cross-entropy loss for training:
\begin{equation}
\mathcal{L} = -\sum_{i=1}^C y_i \log(\hat{y}_i)
\end{equation}
where $C$ is the number of classes, $y_i$ is the true label, and $\hat{y}_i$ is the predicted probability for class $i$.

\paragraph{Optimizer}
We utilize a schedule-free AdamW optimizer \cite{loshchilov2019decoupledweightdecayregularization,defazio2024roadscheduled} .

Further hyperparameters and details can be found in appendix A. 

\section{Results}

This section presents the experimental results from applying the ConvNeXt1D model to the geolocation task using Raman spectra from the RRUFF database. The evaluation focuses on classification accuracy, precision, recall, and F1-score.

\subsection{Dataset Statistics}
After preprocessing, the dataset contained 32,892 across the 101 geographical locations (countries). Table~\ref{tab:dataset_statistics} provides a breakdown of the number of samples per country after filtering. Only countries with at least two samples were retained to ensure meaningful classification results.

\begin{table}[H]
    \centering
    \caption{Sample distribution by country after preprocessing.}
    \label{tab:dataset_statistics}
    \begin{tabular}{lcc}
        \toprule
        \textbf{Country} & \textbf{Sample Count} & \textbf{Percentage} \\
        \midrule
        USA & 9,656 & 30\% \\
        Canada & 2,758 & 16\% \\
        Australia & 836 & 8\% \\
        Others & 20,742 & 46\% \\
        \bottomrule
    \end{tabular}
\end{table}

\subsection{Overall Performance}
The performance of the ConvNeXt1D model was evaluated using 5-fold cross-validation. We report the resulting classification F1-scores and precision on a per-country level (averaged over 5 folds). We also report overall average metrics and their variability across folds.

\subsection{Analysis of Country-Level Performance}
\begin{figure}[H]
    \centering
    \includegraphics[width=\textwidth]{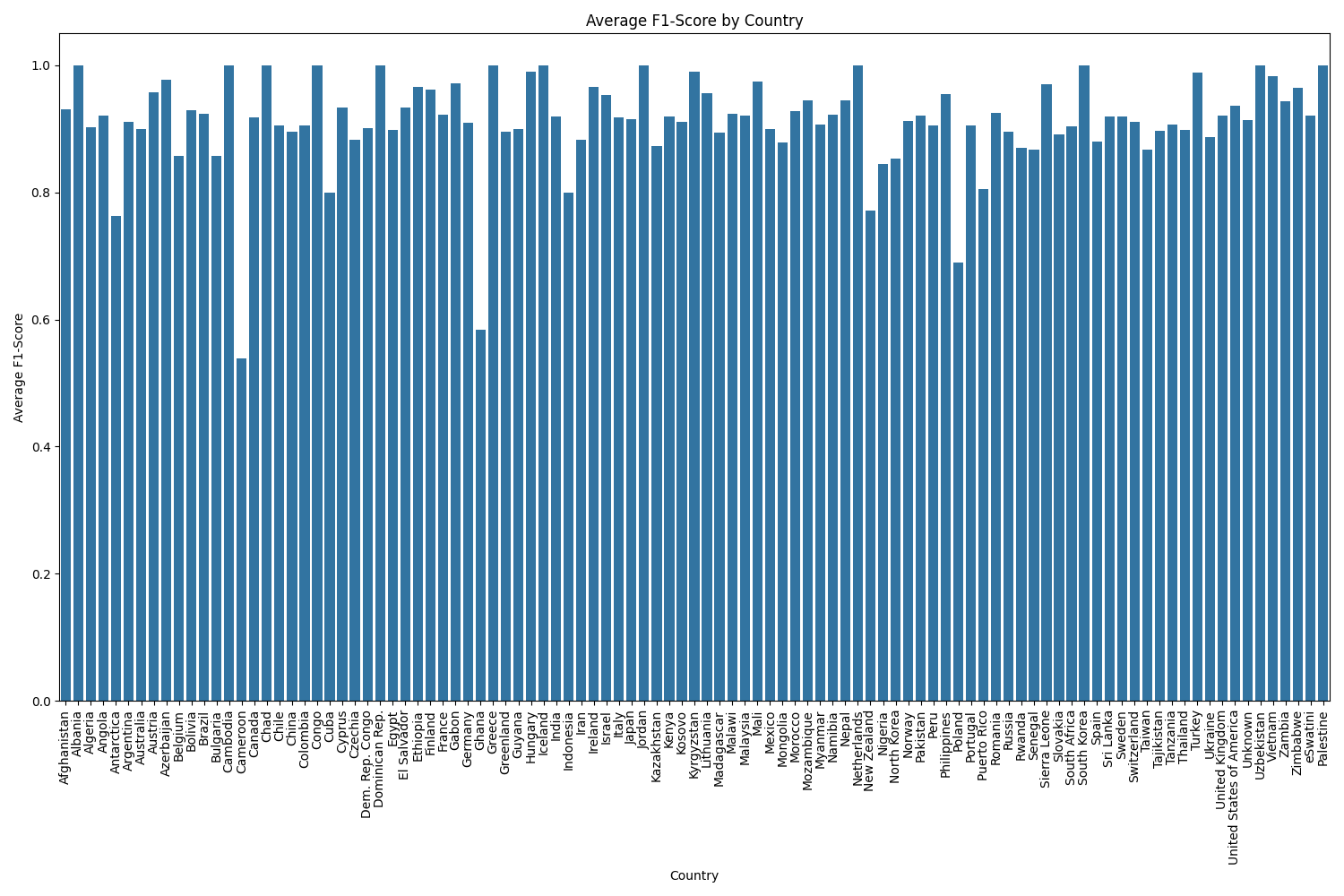}
    \caption{Average F1-score by country.}
    \label{fig:country_F1}
\end{figure}

To analyze model performance, we calculated the average F1-score for each country. Figure~\ref{fig:country_F1} shows the bar plot of the average F1-score by country.

\begin{figure}[H]
    \centering
    \includegraphics[width=\textwidth]{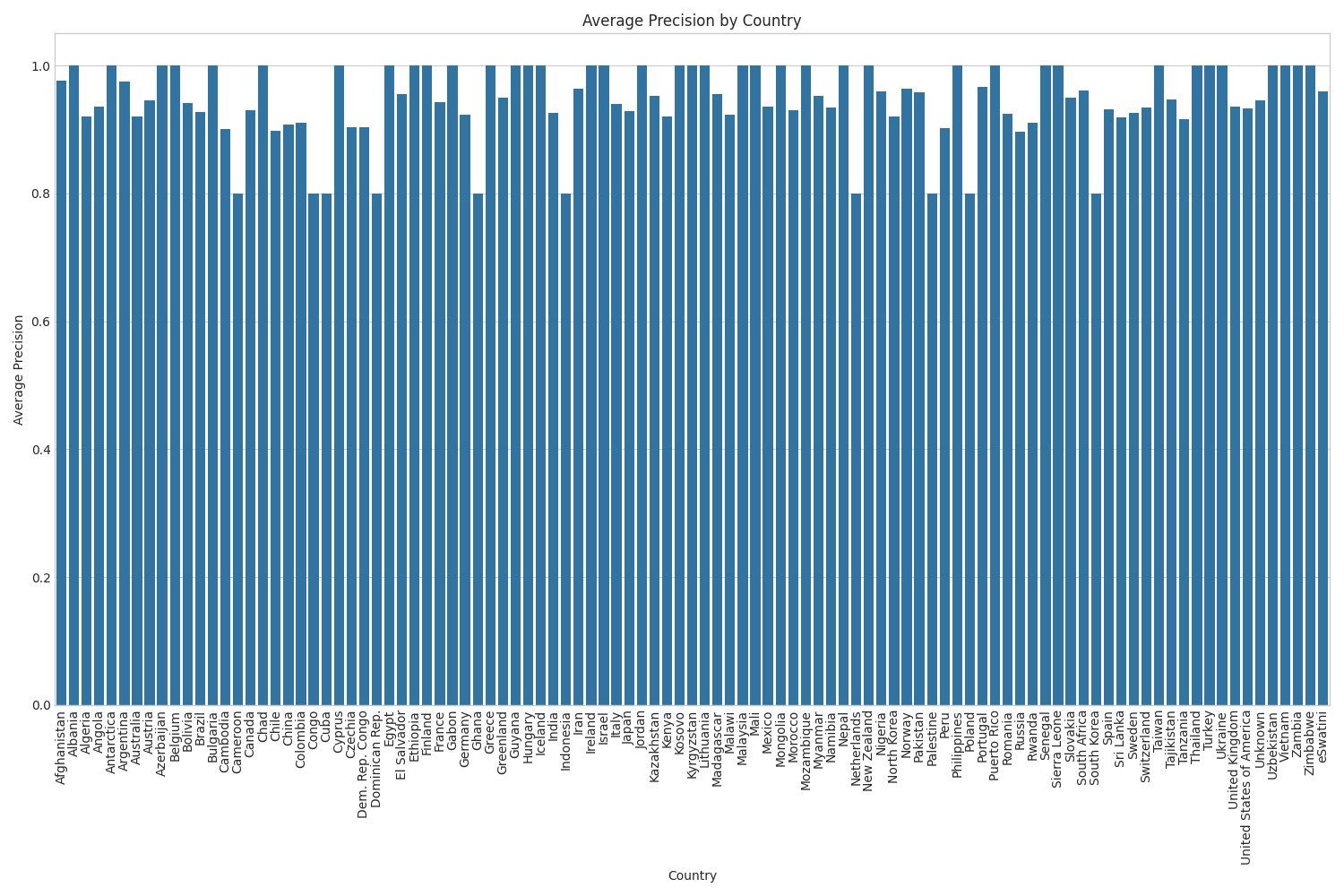}
    \caption{Average precision by country.}
    \label{fig:country_precision}
\end{figure}

\begin{figure}[H]
    \centering
    \includegraphics[width=\textwidth]{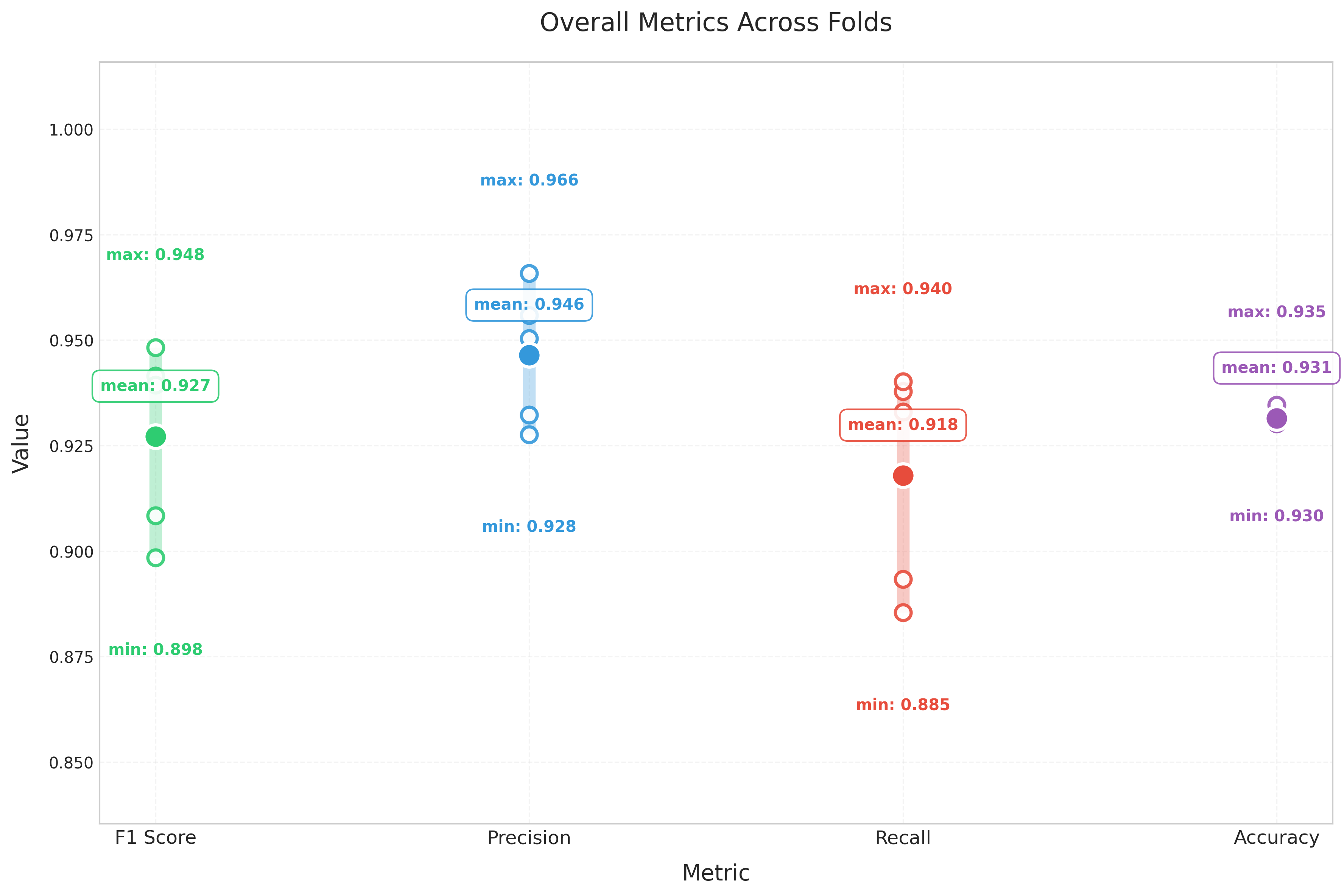}
    \caption{Overall test metrics across 5 folds.}
    \label{fig:overall_metrics_boxplot.png}
\end{figure}

\subsection{Limitations}
Our work is an initial step in the direction of automating geolocation for minerals. Specifically, our dataset, while extremely diverse, has some limiting factors that may affect the results we see in classification of geographical origin. Specifically, the class imbalance shown in the location distribution may play a role in inserting a bias in the trained model. Moreover, specific spurious effects related to sample preparation, as well as to optical effects independent of the mineral's composition, may interfere with a properly generalizable technique for mineral geolocation. Despite the class imbalance, the per country classification metrics shown in \ref{fig:country_F1}\ref{fig:country_precision} seem robust, with little evidence of an important bias affecting geolocation. For this reason, we decided not to employ oversampling techniques, or other methods to mitigated biases due to class imbalances. Future work will explore this aspect in more detail. 

\subsection{Conclusion}
In this work, we introduced an end-to-end deep learning pipeline, centered around the ConvNeXt1D architecture, to process Raman Spectroscopy data, and trained a model to geolocate minerals from their Raman spectra alone, achieving more than a 93\% average accuracy, validating the combined expressive power of vibrational spectroscopy data and deep learning techniques for mineral geolocation. Future work will expand the following results in at least a few directions, among which:
\begin{itemize}
    \item scaling law estimation for representation learning of spectroscopic data
    \item multimodal use of diverse spectroscopic and non-spectroscopic information for representation learning and downstream tasks
    \item extension of our geolocation pipeline to provide a more detailed, robust and accurate geographic prediction of a mineral's origin. 
\end{itemize}

\section*{Conflict of Interest Statement}
The authors declare that there is no conflict of interest regarding the publication of this article.

\section*{Acknowledgments}
We thank the RRUFF Project for providing access to the mineral spectral data.

\bibliographystyle{unsrt}
\bibliography{biblio}
\appendix

\section{Data Preprocessing Pipeline}

This appendix provides a detailed description of the data preprocessing steps employed in the study, including data loading, geocoding of sample localities, handling of missing data, spectral resampling, and normalization. These processes ensure the quality and consistency of the dataset.

\subsection{Geocoding Locality Information}

Each mineral sample in the RRUFF database is associated with a textual locality description that often contains geographic information. To convert these descriptions into precise geographic coordinates (latitude and longitude), we employed multiple geocoding services to enhance coverage and accuracy:

\begin{itemize} \item \textbf{Nominatim} \cite{Nominatim}: An open-source geocoder based on OpenStreetMap data. \item \textbf{Photon} \cite{Photon}: An open-source geocoder built for OpenStreetMap data. \item \textbf{ArcGIS Geocoding Service} \cite{Esri}: A geocoding service provided by Esri. \end{itemize}

The geocoding process involved several steps:

\begin{enumerate} \item \textbf{Data Cleaning}: Locality strings were cleaned by removing text within parentheses, extra spaces, and standardizing delimiters to ensure a consistent input format. \item \textbf{Sequential Geocoding}: Each cleaned locality string was passed to each geocoder in sequence. If a valid geocode was obtained, the process was halted; otherwise, the next service was attempted. \item \textbf{Partial Matching}: For locality descriptions that could not be fully geocoded, partial matches were attempted by progressively truncating parts of the string from the beginning. \end{enumerate}

Despite these efforts, some localities remained un-geocoded due to insufficient or ambiguous information. These samples were flagged and excluded from geographic analyses.

\subsection{Identification of Synthetic Samples}

The dataset includes both natural and synthetic mineral samples. To differentiate these, specific keywords indicative of synthetic origins, such as "synthetic," "laboratory-grown," or "man-made," were identified in the \texttt{locality} and \texttt{names} fields. Samples containing these keywords were classified as synthetic to ensure the model focused on natural mineral occurrences.

\subsection{Handling of Missing Data}

Addressing missing data was essential for maintaining dataset integrity. We applied the following strategies:

\begin{itemize} \item \textbf{Missing Coordinates}: Samples without latitude or longitude values after geocoding were flagged and excluded from analyses requiring geographic information, such as geolocation modeling. \item \textbf{Incomplete Metadata}: Samples with missing essential metadata fields, such as mineral names or spectral data, were reviewed. If critical information was absent and could not be recovered, the sample was removed to prevent potential biases or errors. \end{itemize}

This systematic handling of missing data helped ensure a comprehensive and reliable dataset.

\subsection{Spectral Data Processing}

To prepare spectral data for input into the neural network, we implemented several processing steps:

\paragraph{Global Wavenumber Window Determination}

We identified a global wavenumber window that encompasses the entire range of all spectra. This window standardizes the spectral data, ensuring all spectra are represented over the same wavenumber range.

\paragraph{Spectral Padding}

Spectra were padded with zeros in regions outside their original data to fit the global wavenumber window. This padding ensured all spectra had consistent length.

\paragraph{Normalization}

Min-Max normalization was applied to the intensities of each spectrum, transforming values to a common range (e.g., 0 to 1) and improving training stability.

\paragraph{Interpolation and Resampling}

Spectra were interpolated using cubic splines for smooth representations and resampled onto a fixed-size grid to ensure uniform input dimensions, facilitating batch processing in the neural network.

\section{Training Procedure and Configuration}

In this section, we outline the training procedure for the ConvNeXt1D model, including data splitting, optimization strategies, and model configurations.

\subsection{Data Splitting and Cross-Validation}

To evaluate model performance and ensure generalizability, we employed $k$-fold cross-validation with $k=5$. The dataset was partitioned into five equal subsets (folds) with stratified sampling, preserving the class distribution (countries) across folds. In each iteration, one fold was used as the validation set, and the remaining four folds served as the training set, ensuring each fold was used once for validation. This process reduces overfitting and provides an estimate of model generalization.

\subsection{Model Configuration}

The ConvNeXt1D model was adapted from the ConvNeXt architecture to handle one-dimensional spectral data. It consists of the following core components:

\begin{itemize} \item \textbf{Stem Layer}: An initial convolutional layer with a large kernel size and stride for downsampling. \item \textbf{ConvNeXt1D Blocks}: Stacked blocks that incorporate convolutional and normalization layers. \item \textbf{Classification Head}: A global average pooling layer followed by a fully connected layer mapping features to the number of classes (countries). \end{itemize}

Key hyperparameters for the model were selected based on standard configurations for ConvNeXt, including depth and dimension settings specific to one-dimensional data.

\subsection{Optimization Strategy and Regularization}

The training used the following methods:

\begin{itemize} \item \textbf{Loss Function}: Cross-entropy loss, a standard choice for multi-class classification. \item \textbf{Optimizer}: A scheduler-free AdamW with weight decay for robust convergence. \item \textbf{Gradient Clipping}: Gradient clipping by norm to ensure stable training and prevent exploding gradients. \end{itemize}

To further improve generalization, we incorporated additional regularization techniques:

\begin{itemize} \item \textbf{Dropout}: Preventing feature co-adaptation in network layers. \item \textbf{Stochastic Depth (DropPath)}: Randomly dropping layers during training, which encourages model robustness. \item \textbf{Weight Decay}: Included in AdamW for regularizing large weights, with a value of 0.35 (due to overfitting tendency of the networks). \end{itemize}
\end{document}